\documentclass[a4paper, 10pt, conference]{ieeeconf}      % Use this line for a4 paper

\IEEEoverridecommandlockouts                              % This command is only needed if 
\usepackage{placeins}
\usepackage{graphicx} % for pdf, bitmapped graphics files
\usepackage{amsmath} % assumes amsmath package installed
\usepackage[caption=false]{subfig}

\usepackage{url}
\usepackage{siunitx}
\usepackage{tabularx, booktabs}
\usepackage{multirow}
\usepackage{lipsum}
\usepackage[ruled,vlined]{algorithm2e}
\usepackage{amsfonts}
\usepackage{xcolor}
\usepackage{pgfplots}

\let\labelindent\relax
\usepackage{enumitem}
\usepackage{textcomp}

\usepackage{tikz}
\usetikzlibrary{arrows}

\newcommand{\comment}[1]{}

\title{\LARGE \bf
siaNMS: Non-Maximum Suppression with Siamese Networks for Multi-Camera 3D Object Detection
}
\author{Irene Cort\'{e}s, Jorge Beltr\'{a}n, Arturo de la Escalera and Fernando Garc\'{i}a\\ Intelligent Systems Laboratory (LSI) Research Group\\
Universidad Carlos III de Madrid, Legan\'{e}s, Madrid, Spain\\
\{irecorte, jbeltran, escalera, fegarcia\}@ing.uc3m.es 
}

\begin{document}

\maketitle

\thispagestyle{empty}
\pagestyle{empty}

%%%%%%%%%%%%%%%%%%%%%%%%%%%%%%%%%%%%%%%%%%%%%%%%%%%%%%%%%%%%%%%%%%%%%%%%%%%%%%%%
\begin{abstract}
 %The rapid development of 3D object detection for autonomous vehicles brings the possibility to handle more complex scenarios. Thus, 
The rapid development of embedded hardware in autonomous vehicles broadens their computational capabilities, thus bringing the possibility to mount more complete sensor setups able to handle driving scenarios of higher complexity. As a result, new challenges such as multiple detections of the same object have to be addressed. In this work, a siamese network is integrated into the pipeline of a well-known 3D object detector approach to suppress duplicate proposals coming from different cameras via re-identification. Additionally, associations are exploited to enhance the 3D box regression of the object by aggregating their corresponding LiDAR frustums. The experimental evaluation on the nuScenes dataset shows that the proposed method outperforms traditional NMS approaches.
\end{abstract}

%%%%%%%%%%%%%%%%%%%%%%%%%%%%%%%%%%%%%%%%%%%%%%%%%%%%%%%%%%%%%%%%%%%%%%%%%%%%%%%%
\section{INTRODUCTION}

\global\csname @topnum\endcsname 0
\global\csname @botnum\endcsname 0

\begin{figure}
\centering
%\missingfigure{Figure is missing}
\includegraphics[width=\linewidth]{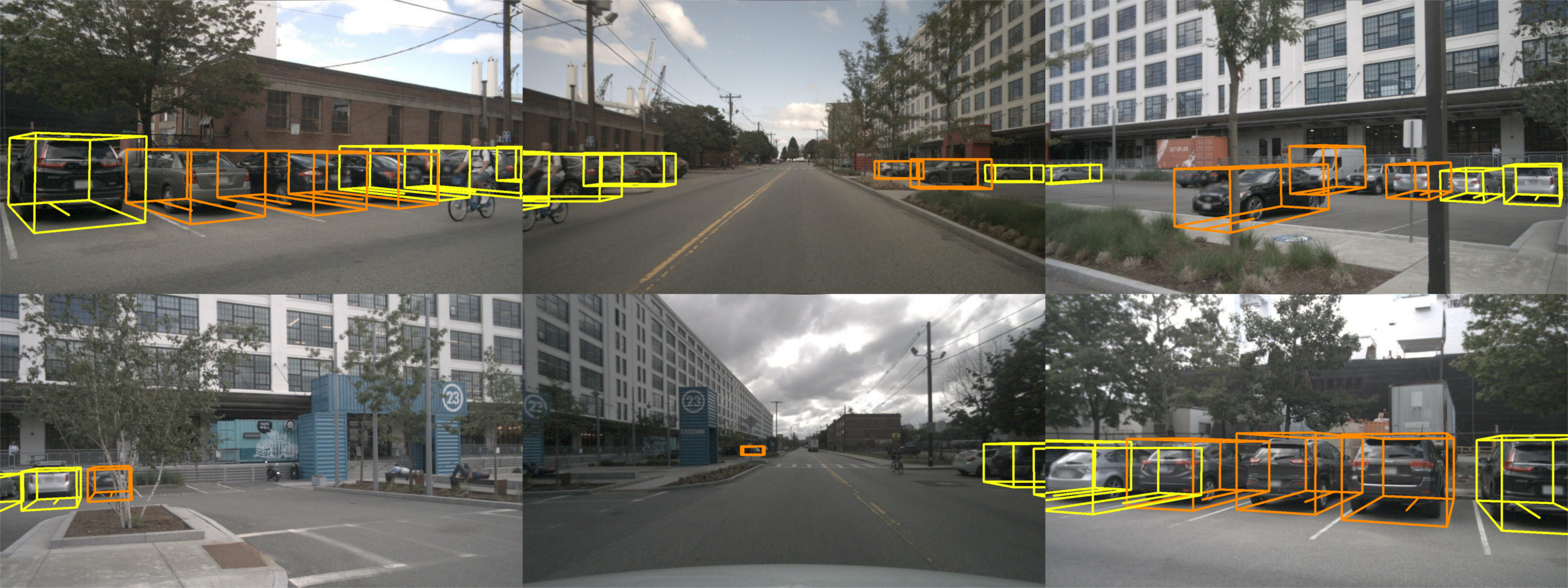}
\caption{Sample of NuScenes labels. Objects on a single image are colored in orange, while those on two consecutive cameras are shown in yellow.}
%\caption{Ground truth detection on the NuScenes dataset. Single image detections are colored in red. Meanwhile, detection shared between two cameras are colored blue.}
\vspace{-1em}
\label{fig:nuscenes_gt}
\end{figure}

Self-driving vehicles are going to change the future of transportation in terms of safety, efficiency, and pollution. One of the key advantages of this upcoming technology is the potential to reduce the number of road accidents. For this reason, vehicle control will rely on an accurate perception of the environment.

The task of 3D object detection plays a major role in the evolution of the autonomous driving field. However, since supervised deep neural networks have outperformed any pre-existing methods, its development highly depends on the availability of public annotated datasets. In this regard, KITTI \cite{kitti_object} has been considered the standard benchmark for many years now. Nonetheless, despite the quality of the annotations and the variety of scenes, it has proved to be insufficient to build robust object detectors for driving scenarios with adverse light or weather conditions.

To tackle this lack of data availability and boost research, many datasets have been released recently. Unlike KITTI, these new collections usually include more complete sensor setups with higher redundancy and are tailored for \ang{360} perception. Thus, we can find nuScenes \cite{nuscenes},  Waymo \cite{waymo}, Lyft \cite{lyft2019}, or Argoverse \cite{argoverse}, which include information captured from several surrounding cameras, one or more laser scanners, radars, GPS, and whose sizes are several orders of magnitude greater than KITTI's.

% Oportunity to handle more complex scenarios, and thus enabling a fully autonomous control of the cars.
The emergence of these new annotated datasets is a big opportunity for progress in the 3D object detection domain, but at the same time, it opens up a set of new challenging tasks, such as processing a vast quantity of data in real-time, or dealing with misalignments among detections coming from different sources of information. Although the first issue may be solved by scaling the hardware, the latter especially affects those methods that take images as input, as they can only work with a limited view of the environment.

% TODO Quizás explicar con una línea más lo que quiere darse a entender en el párrafo anterior

In this paper, we insert a re-identification module in a popular state-of-the-art 3D object detector, F-PointNets \cite{qi2018frustum}, to improve the performance of the box regression of objects on the side of the image. The proposed framework is fed with pairs of 2D proposals from contiguous surrounding cameras mounted on a \ang{360} on-board setup, and provides a similarity estimation so that detections of the same object in different cameras can be matched. In this manner, the siamese network permits not only to suppress multiple detections of the same obstacle in a traditional Non-Maximum Suppresion (NMS) fashion, but also to aggregate the corresponding LiDAR points associated with both occurrences. As a result, a more faithful and complete representation of the object in the spatial modality is created, which can enhance the 3D box estimation performed in the last step of the pipeline.

The rest of this paper is structured as follows. In Section II, a review of the related works is provided. Sections III and IV include the description of the proposed approach and the details of the network design and training, respectively. The experimental results are discussed in Section V. Finally, conclusions and future work are presented in Section VI.

%%%%%%%%%%%%%%%%%%%%%%%%%%%%%%%%%%%%%%%%%%%%%%%%%%%%%%%%%%%%%%%%%%%%%%%%%%%%%%%%
\section{RELATED WORK}

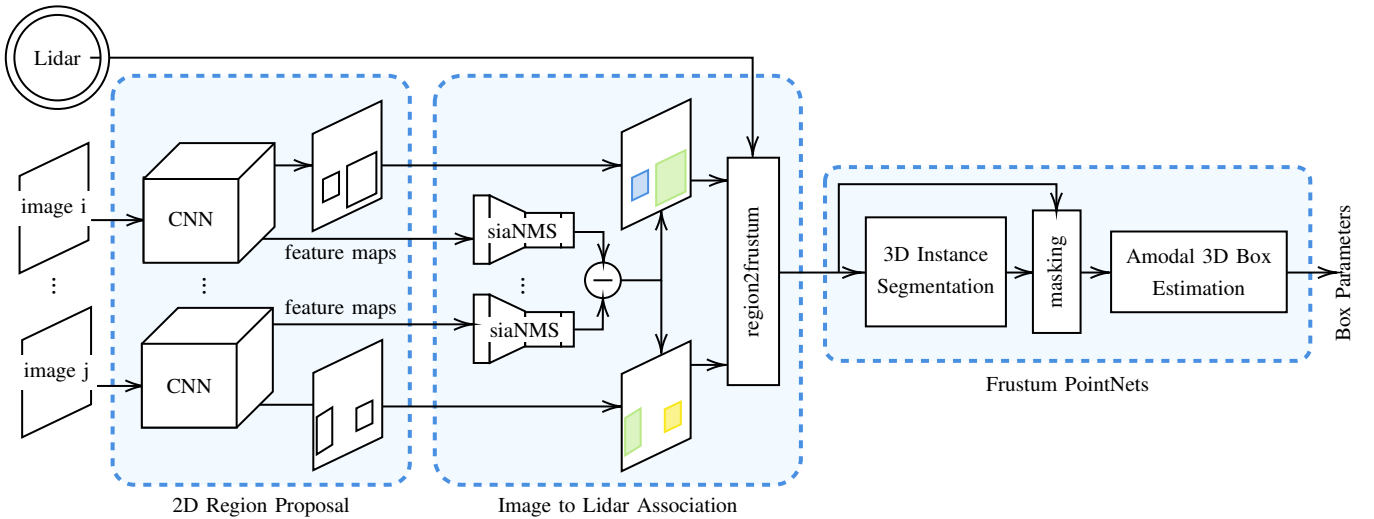
\begin{figure*}[ht]
\tikzset{every picture/.style={line width=0.75pt}} %set default line width to 0.75pt        
\begin{tikzpicture}[x=0.645pt,y=0.645pt,yscale=-1,xscale=1]
%uncomment if require: \path (0,732); %set diagram left start at 0, and has height of 732

%Rounded Rect [id:dp9548233425392134] 
\draw  [color={rgb, 255:red, 74; green, 144; blue, 226 }  ,draw opacity=1 ][fill={rgb, 255:red, 231; green, 244; blue, 255 }  ,fill opacity=0.54 ][line width=1.5][dashed]  (576,270.29) .. controls (576,265.71) and (579.71,262) .. (584.29,262) -- (849.37,262) .. controls (853.95,262) and (857.67,265.71) .. (857.67,270.29) -- (857.67,367.04) .. controls (857.67,371.62) and (853.95,375.33) .. (849.37,375.33) -- (584.29,375.33) .. controls (579.71,375.33) and (576,371.62) .. (576,367.04) -- cycle ;
%Rounded Rect [id:dp11824443588603173] 
\draw  [color={rgb, 255:red, 74; green, 144; blue, 226 }  ,draw opacity=1 ][fill={rgb, 255:red, 231; green, 244; blue, 255 }  ,fill opacity=0.54 ][line width=1.5][dashed]  (349.5,224.15) .. controls (349.5,215.56) and (356.46,208.6) .. (365.05,208.6) -- (546.45,208.6) .. controls (555.04,208.6) and (562,215.56) .. (562,224.15) -- (562,432.25) .. controls (562,440.84) and (555.04,447.8) .. (546.45,447.8) -- (365.05,447.8) .. controls (356.46,447.8) and (349.5,440.84) .. (349.5,432.25) -- cycle ;
%Rounded Rect [id:dp3800388069285543] 
\draw  [color={rgb, 255:red, 74; green, 144; blue, 226 }  ,draw opacity=1 ][fill={rgb, 255:red, 231; green, 244; blue, 255 }  ,fill opacity=0.54 ][line width=1.5][dashed]  (162.5,221.22) .. controls (162.5,214.25) and (168.15,208.6) .. (175.12,208.6) -- (322.38,208.6) .. controls (329.35,208.6) and (335,214.25) .. (335,221.22) -- (335,435.18) .. controls (335,442.15) and (329.35,447.8) .. (322.38,447.8) -- (175.12,447.8) .. controls (168.15,447.8) and (162.5,442.15) .. (162.5,435.18) -- cycle ;
%Straight Lines [id:da5131337180237594] 
\draw    (242,401) -- (307,401) ;
\draw [shift={(309,401)}, rotate = 180] [color={rgb, 255:red, 0; green, 0; blue, 0 }  ][line width=0.75]    (10.93,-3.29) .. controls (6.95,-1.4) and (3.31,-0.3) .. (0,0) .. controls (3.31,0.3) and (6.95,1.4) .. (10.93,3.29)   ;

%Flowchart: Data [id:dp3574736787929653] 
\draw   (148.5,247.89) -- (148.5,304.77) -- (108.5,324.04) -- (108.5,267.16) -- cycle ;
%Shape: Rectangle [id:dp6699384948420253] 
\draw  [color={rgb, 255:red, 0; green, 0; blue, 0 }  ,draw opacity=0 ][fill={rgb, 255:red, 255; green, 255; blue, 255 }  ,fill opacity=1 ] (105,275.96) -- (152,275.96) -- (152,295.96) -- (105,295.96) -- cycle ;
%Flowchart: Data [id:dp6737368174843525] 
\draw   (150,343.89) -- (150,400.77) -- (110,420.04) -- (110,363.16) -- cycle ;
%Straight Lines [id:da39539929140153185] 
\draw    (150,293) -- (179.5,293) ;
\draw [shift={(181.5,293)}, rotate = 180] [color={rgb, 255:red, 0; green, 0; blue, 0 }  ][line width=0.75]    (10.93,-3.29) .. controls (6.95,-1.4) and (3.31,-0.3) .. (0,0) .. controls (3.31,0.3) and (6.95,1.4) .. (10.93,3.29)   ;

%Straight Lines [id:da65563380502881] 
\draw    (226.5,261) -- (276.67,261) ;
\draw [shift={(278.67,261)}, rotate = 180] [color={rgb, 255:red, 0; green, 0; blue, 0 }  ][line width=0.75]    (10.93,-3.29) .. controls (6.95,-1.4) and (3.31,-0.3) .. (0,0) .. controls (3.31,0.3) and (6.95,1.4) .. (10.93,3.29)   ;

%Straight Lines [id:da3747780673960932] 
\draw    (150,390) -- (178.85,390) ;
\draw [shift={(180.85,390)}, rotate = 180] [color={rgb, 255:red, 0; green, 0; blue, 0 }  ][line width=0.75]    (10.93,-3.29) .. controls (6.95,-1.4) and (3.31,-0.3) .. (0,0) .. controls (3.31,0.3) and (6.95,1.4) .. (10.93,3.29)   ;

%Shape: Rectangle [id:dp27906194852327637] 
\draw  [fill={rgb, 255:red, 255; green, 255; blue, 255 }  ,fill opacity=1 ] (372.11,336.56) -- (381.36,336.56) -- (381.36,376.56) -- (372.11,376.56) -- cycle ;
%Shape: Trapezoid [id:dp1118513001570538] 
\draw  [fill={rgb, 255:red, 255; green, 255; blue, 255 }  ,fill opacity=1 ] (381.36,336.56) -- (402,346.4) -- (402,366.73) -- (381.36,376.56) -- cycle ;
%Shape: Rectangle [id:dp4453145460335879] 
\draw  [fill={rgb, 255:red, 255; green, 255; blue, 255 }  ,fill opacity=1 ] (402,346.8) -- (422.8,346.8) -- (422.8,366.98) -- (402,366.98) -- cycle ;
%Shape: Rectangle [id:dp7639368386935612] 
\draw  [fill={rgb, 255:red, 255; green, 255; blue, 255 }  ,fill opacity=1 ] (422.8,346.8) -- (430.19,346.8) -- (430.19,366.98) -- (422.8,366.98) -- cycle ;
%Shape: Rectangle [id:dp8062254404679752] 
\draw  [color={rgb, 255:red, 0; green, 0; blue, 0 }  ,draw opacity=0 ][fill={rgb, 255:red, 255; green, 255; blue, 255 }  ,fill opacity=1 ] (374.4,349.8) -- (429,349.8) -- (429,364.2) -- (374.4,364.2) -- cycle ;

%Straight Lines [id:da6757643285338268] 
\draw    (217.58,353.96) -- (369.57,353.96) ;
\draw [shift={(371.57,353.96)}, rotate = 180] [color={rgb, 255:red, 0; green, 0; blue, 0 }  ][line width=0.75]    (10.93,-3.29) .. controls (6.95,-1.4) and (3.31,-0.3) .. (0,0) .. controls (3.31,0.3) and (6.95,1.4) .. (10.93,3.29)   ;

%Straight Lines [id:da1179793825769575] 
\draw    (208.2,303.96) -- (370.04,303.96) ;
\draw [shift={(372.04,303.96)}, rotate = 180] [color={rgb, 255:red, 0; green, 0; blue, 0 }  ][line width=0.75]    (10.93,-3.29) .. controls (6.95,-1.4) and (3.31,-0.3) .. (0,0) .. controls (3.31,0.3) and (6.95,1.4) .. (10.93,3.29)   ;

%Shape: Circle [id:dp7626975515851726] 
\draw   [fill={rgb, 255:red, 255; green, 255; blue, 255 }  ,fill opacity=1 ](436.7,328.2) .. controls (436.7,322.46) and (441.36,317.8) .. (447.1,317.8) .. controls (452.84,317.8) and (457.5,322.46) .. (457.5,328.2) .. controls (457.5,333.94) and (452.84,338.6) .. (447.1,338.6) .. controls (441.36,338.6) and (436.7,333.94) .. (436.7,328.2) -- cycle ;
%Straight Lines [id:da184241471803819] 
\draw    (441.6,328.2) -- (452.6,328.2) ;

%Straight Lines [id:da05496093945220992] 
\draw    (430.65,299.96) -- (447.4,299.96) -- (447.4,315.8) ;
\draw [shift={(447.4,317.8)}, rotate = 270] [color={rgb, 255:red, 0; green, 0; blue, 0 }  ][line width=0.75]    (10.93,-3.29) .. controls (6.95,-1.4) and (3.31,-0.3) .. (0,0) .. controls (3.31,0.3) and (6.95,1.4) .. (10.93,3.29)   ;

%Straight Lines [id:da2159875308224124] 
\draw    (430.5,356.96) -- (447,356.96) -- (447.09,340.6) ;
\draw [shift={(447.1,338.6)}, rotate = 450.31] [color={rgb, 255:red, 0; green, 0; blue, 0 }  ][line width=0.75]    (10.93,-3.29) .. controls (6.95,-1.4) and (3.31,-0.3) .. (0,0) .. controls (3.31,0.3) and (6.95,1.4) .. (10.93,3.29)   ;

%Shape: Rectangle [id:dp07610566604122582] 
\draw  [fill={rgb, 255:red, 255; green, 255; blue, 255 }  ,fill opacity=1 ] (372.11,278.56) -- (381.36,278.56) -- (381.36,318.56) -- (372.11,318.56) -- cycle ;
%Shape: Trapezoid [id:dp13158067534273932] 
\draw  [fill={rgb, 255:red, 255; green, 255; blue, 255 }  ,fill opacity=1 ] (381.36,278.56) -- (402,288.4) -- (402,308.73) -- (381.36,318.56) -- cycle ;
%Shape: Rectangle [id:dp49253956077865] 
\draw  [fill={rgb, 255:red, 255; green, 255; blue, 255 }  ,fill opacity=1 ] (402,288.8) -- (422.8,288.8) -- (422.8,308.98) -- (402,308.98) -- cycle ;
%Shape: Rectangle [id:dp6635253384678481] 
\draw  [fill={rgb, 255:red, 255; green, 255; blue, 255 }  ,fill opacity=1 ] (422.8,288.8) -- (430.19,288.8) -- (430.19,308.98) -- (422.8,308.98) -- cycle ;
%Shape: Rectangle [id:dp501392654995769] 
\draw  [color={rgb, 255:red, 0; green, 0; blue, 0 }  ,draw opacity=0 ][fill={rgb, 255:red, 255; green, 255; blue, 255 }  ,fill opacity=1 ] (373.2,292) -- (427.8,292) -- (427.8,306.4) -- (373.2,306.4) -- cycle ;

%Straight Lines [id:da870815220308433] 
\draw    (487,269.84) -- (517.67,269.84) ;
\draw [shift={(519.67,269.84)}, rotate = 180] [color={rgb, 255:red, 0; green, 0; blue, 0 }  ][line width=0.75]    (10.93,-3.29) .. controls (6.95,-1.4) and (3.31,-0.3) .. (0,0) .. controls (3.31,0.3) and (6.95,1.4) .. (10.93,3.29)   ;

%Straight Lines [id:da729109835346244] 
\draw    (457.5,328.2) -- (481,328.2) -- (481,290) ;
\draw [shift={(481,288)}, rotate = 450] [color={rgb, 255:red, 0; green, 0; blue, 0 }  ][line width=0.75]    (10.93,-3.29) .. controls (6.95,-1.4) and (3.31,-0.3) .. (0,0) .. controls (3.31,0.3) and (6.95,1.4) .. (10.93,3.29)   ;

%Straight Lines [id:da6098691159392942] 
\draw    (481,328.2) -- (481,368.6) ;
\draw [shift={(481,370.6)}, rotate = 270] [color={rgb, 255:red, 0; green, 0; blue, 0 }  ][line width=0.75]    (10.93,-3.29) .. controls (6.95,-1.4) and (3.31,-0.3) .. (0,0) .. controls (3.31,0.3) and (6.95,1.4) .. (10.93,3.29)   ;

%Straight Lines [id:da16390646534783704] 
\draw    (310,402) -- (455.67,402) ;
\draw [shift={(457.67,402)}, rotate = 180] [color={rgb, 255:red, 0; green, 0; blue, 0 }  ][line width=0.75]    (10.93,-3.29) .. controls (6.95,-1.4) and (3.31,-0.3) .. (0,0) .. controls (3.31,0.3) and (6.95,1.4) .. (10.93,3.29)   ;

%Straight Lines [id:da22041180643215275] 
\draw    (310.78,261) -- (455.67,261) ;
\draw [shift={(457.67,261)}, rotate = 180] [color={rgb, 255:red, 0; green, 0; blue, 0 }  ][line width=0.75]    (10.93,-3.29) .. controls (6.95,-1.4) and (3.31,-0.3) .. (0,0) .. controls (3.31,0.3) and (6.95,1.4) .. (10.93,3.29)   ;

%Straight Lines [id:da5977075519621775] 
\draw    (487.67,377.69) -- (516.33,377.69) ;
\draw [shift={(518.33,377.69)}, rotate = 180] [color={rgb, 255:red, 0; green, 0; blue, 0 }  ][line width=0.75]    (10.93,-3.29) .. controls (6.95,-1.4) and (3.31,-0.3) .. (0,0) .. controls (3.31,0.3) and (6.95,1.4) .. (10.93,3.29)   ;

%Straight Lines [id:da3210328594408123] 
\draw    (149.5,198) -- (534,198) -- (534,253) ;
\draw [shift={(534,255)}, rotate = 270] [color={rgb, 255:red, 0; green, 0; blue, 0 }  ][line width=0.75]    (10.93,-3.29) .. controls (6.95,-1.4) and (3.31,-0.3) .. (0,0) .. controls (3.31,0.3) and (6.95,1.4) .. (10.93,3.29)   ;

%Shape: Rectangle [id:dp35008226064245473] 
\draw   (180.5,268) -- (235.5,268) -- (235.5,317) -- (180.5,317) -- cycle ;

%Shape: Rectangle [id:dp29077984056513095] 
\draw   (179.5,365.33) -- (234.5,365.33) -- (234.5,414.33) -- (179.5,414.33) -- cycle ;

%Shape: Cube [id:dp5925591788516786] 
\draw  [fill={rgb, 255:red, 255; green, 255; blue, 255 }  ,fill opacity=1 ] (179.5,365.33) -- (200.5,344.33) -- (255.5,344.33) -- (255.5,393.33) -- (234.5,414.33) -- (179.5,414.33) -- cycle ; \draw   (255.5,344.33) -- (234.5,365.33) -- (179.5,365.33) ; \draw   (234.5,365.33) -- (234.5,414.33) ;
%Flowchart: Data [id:dp6713390837372748] 
\draw  [fill={rgb, 255:red, 255; green, 255; blue, 255 }  ,fill opacity=1 ] (318.6,223.29) -- (318.6,280.17) -- (278.6,299.44) -- (278.6,242.56) -- cycle ;
%Flowchart: Data [id:dp843356171165581] 
\draw   (315.6,253.18) -- (315.6,277.35) -- (298.6,285.54) -- (298.6,261.36) -- cycle ;
%Flowchart: Data [id:dp9683519061055057] 
\draw   (293.92,264.77) -- (293.92,278.02) -- (284.6,282.5) -- (284.6,269.26) -- cycle ;

%Flowchart: Data [id:dp02041449621076241] 
\draw  [fill={rgb, 255:red, 255; green, 255; blue, 255 }  ,fill opacity=1 ] (319,363.09) -- (319,419.97) -- (279,439.24) -- (279,382.36) -- cycle ;
%Flowchart: Data [id:dp8232643593728104] 
\draw   (290,402.53) -- (290,424.36) -- (281,430.03) -- (281,408.2) -- cycle ;
%Flowchart: Data [id:dp21456621637700546] 
\draw   (313.32,398.57) -- (313.32,411.82) -- (304,416.3) -- (304,403.06) -- cycle ;

%Shape: Cube [id:dp17340659621120058] 
\draw  [fill={rgb, 255:red, 255; green, 255; blue, 255 }  ,fill opacity=1 ] (180.5,268.96) -- (201.5,247.96) -- (256.5,247.96) -- (256.5,296.96) -- (235.5,317.96) -- (180.5,317.96) -- cycle ; \draw   (256.5,247.96) -- (235.5,268.96) -- (180.5,268.96) ; \draw   (235.5,268.96) -- (235.5,317.96) ;
%Straight Lines [id:da5021537970064853] 
\draw    (538.33,323.67) -- (595,323.67) ;
\draw [shift={(597,323.67)}, rotate = 180] [color={rgb, 255:red, 0; green, 0; blue, 0 }  ][line width=0.75]    (10.93,-3.29) .. controls (6.95,-1.4) and (3.31,-0.3) .. (0,0) .. controls (3.31,0.3) and (6.95,1.4) .. (10.93,3.29)   ;

%Straight Lines [id:da13650672750422177] 
\draw    (597,323.67) -- (695,323.67) ;
\draw [shift={(697,323.67)}, rotate = 180] [color={rgb, 255:red, 0; green, 0; blue, 0 }  ][line width=0.75]    (10.93,-3.29) .. controls (6.95,-1.4) and (3.31,-0.3) .. (0,0) .. controls (3.31,0.3) and (6.95,1.4) .. (10.93,3.29)   ;

%Straight Lines [id:da787089983483432] 
\draw    (710.5,323.67) -- (739,323.67) ;
\draw [shift={(741,323.67)}, rotate = 180] [color={rgb, 255:red, 0; green, 0; blue, 0 }  ][line width=0.75]    (10.93,-3.29) .. controls (6.95,-1.4) and (3.31,-0.3) .. (0,0) .. controls (3.31,0.3) and (6.95,1.4) .. (10.93,3.29)   ;

%Shape: Rectangle [id:dp13760984861199987] 
\draw  [fill={rgb, 255:red, 255; green, 255; blue, 255 }  ,fill opacity=1 ] (696.67,287.33) -- (724.33,287.33) -- (724.33,359) -- (696.67,359) -- cycle ;

%Shape: Rectangle [id:dp784496498201628] 
\draw  [fill={rgb, 255:red, 255; green, 255; blue, 255 }  ,fill opacity=1 ] (599.69,291.38) -- (681,291.38) -- (681,355.17) -- (599.69,355.17) -- cycle ;

%Straight Lines [id:da6572359414358393] 
\draw    (584,323.5) -- (584,272) -- (710.33,272) -- (710.33,286) ;
\draw [shift={(710.33,288)}, rotate = 270] [color={rgb, 255:red, 0; green, 0; blue, 0 }  ][line width=0.75]    (10.93,-3.29) .. controls (6.95,-1.4) and (3.31,-0.3) .. (0,0) .. controls (3.31,0.3) and (6.95,1.4) .. (10.93,3.29)   ;

%Straight Lines [id:da8947687359251715] 
\draw    (771.19,323.67) -- (871,323.67) ;
\draw [shift={(873,323.67)}, rotate = 180] [color={rgb, 255:red, 0; green, 0; blue, 0 }  ][line width=0.75]    (10.93,-3.29) .. controls (6.95,-1.4) and (3.31,-0.3) .. (0,0) .. controls (3.31,0.3) and (6.95,1.4) .. (10.93,3.29)   ;

%Shape: Rectangle [id:dp9044999381549421] 
\draw  [fill={rgb, 255:red, 255; green, 255; blue, 255 }  ,fill opacity=1 ] (742.05,299.67) -- (844,299.67) -- (844,346.88) -- (742.05,346.88) -- cycle ;

%Shape: Rectangle [id:dp2555788848500111] 
\draw  [color={rgb, 255:red, 0; green, 0; blue, 0 }  ,draw opacity=0 ][fill={rgb, 255:red, 255; green, 255; blue, 255 }  ,fill opacity=1 ] (106.5,371.96) -- (153.5,371.96) -- (153.5,391.96) -- (106.5,391.96) -- cycle ;
%Flowchart: Data [id:dp7812297610935175] 
\draw  [fill={rgb, 255:red, 255; green, 255; blue, 255 }  ,fill opacity=1 ] (498,222.29) -- (498,279.17) -- (458,298.44) -- (458,241.56) -- cycle ;
%Flowchart: Data [id:dp13227931254262892] 
\draw  [color={rgb, 255:red, 184; green, 233; blue, 134 }  ,draw opacity=1 ][fill={rgb, 255:red, 184; green, 233; blue, 134 }  ,fill opacity=0.5 ] (495,252.18) -- (495,276.35) -- (478,284.54) -- (478,260.36) -- cycle ;
%Flowchart: Data [id:dp46735531907922345] 
\draw  [color={rgb, 255:red, 74; green, 144; blue, 226 }  ,draw opacity=1 ][fill={rgb, 255:red, 74; green, 144; blue, 226 }  ,fill opacity=0.3 ] (473.32,263.77) -- (473.32,277.02) -- (464,281.5) -- (464,268.26) -- cycle ;

%Flowchart: Data [id:dp5966124040096517] 
\draw  [fill={rgb, 255:red, 255; green, 255; blue, 255 }  ,fill opacity=1 ] (498,363.69) -- (498,420.57) -- (458,439.84) -- (458,382.96) -- cycle ;
%Flowchart: Data [id:dp6112475144822782] 
\draw  [color={rgb, 255:red, 184; green, 233; blue, 134 }  ,draw opacity=1 ][fill={rgb, 255:red, 184; green, 233; blue, 134 }  ,fill opacity=0.5 ] (469,403.13) -- (469,424.96) -- (460,430.63) -- (460,408.8) -- cycle ;
%Flowchart: Data [id:dp1642524221921393] 
\draw  [color={rgb, 255:red, 248; green, 231; blue, 28 }  ,draw opacity=1 ][fill={rgb, 255:red, 248; green, 231; blue, 28 }  ,fill opacity=0.5 ] (492.32,399.17) -- (492.32,412.42) -- (483,416.9) -- (483,403.66) -- cycle ;

%Shape: Rectangle [id:dp3613854263369345] 
\draw  [fill={rgb, 255:red, 255; green, 255; blue, 255 }  ,fill opacity=1 ] (519.5,256.5) -- (549.33,256.5) -- (549.33,389.5) -- (519.5,389.5) -- cycle ;

%Shape: Donut [id:dp5128185879251252] 
\draw  [fill={rgb, 255:red, 255; green, 255; blue, 255 }  ,fill opacity=1 ,even odd rule] (105.5,198) .. controls (105.5,184.19) and (116.69,173) .. (130.5,173) .. controls (144.31,173) and (155.5,184.19) .. (155.5,198) .. controls (155.5,211.81) and (144.31,223) .. (130.5,223) .. controls (116.69,223) and (105.5,211.81) .. (105.5,198)(100.5,198) .. controls (100.5,181.43) and (113.93,168) .. (130.5,168) .. controls (147.07,168) and (160.5,181.43) .. (160.5,198) .. controls (160.5,214.57) and (147.07,228) .. (130.5,228) .. controls (113.93,228) and (100.5,214.57) .. (100.5,198) ;

\draw (215.67,331.33) node  [rotate=-90] [align=center] {...};
\draw (130.5,198) node   [align=center] {{\footnotesize Lidar}};
\draw (208,292.5) node   [align=center] {{\footnotesize CNN}};
\draw (207,389.83) node   [align=center] {{\footnotesize CNN}};
\draw (130,381.96) node   [align=center] {{\footnotesize image j}};
\draw (534.42,323) node  [rotate=-270] [align=center] {{\footnotesize region2frustum}};
\draw (400.5,299.2) node   [align=center] {{\footnotesize siaNMS}};
\draw (401.7,357) node   [align=center] {{\footnotesize siaNMS}};
\draw (400.67,331.33) node  [rotate=-90] [align=center] {...};
\draw (128.5,331.33) node  [rotate=-90] [align=center] {...};
\draw (793.03,323.28) node   [align=center] {\footnotesize Amodal 3D Box\\ \footnotesize Estimation};
\draw (710.5,323.17) node  [rotate=-270] [align=center] {{\footnotesize masking}};
\draw (640.34,323.27) node   [align=center] {{\footnotesize  3D Instance}\\{\footnotesize Segmentation}};
\draw (876.67,323.94) node  [rotate=-270] [align=center] {{\footnotesize Box Parameters}};
\draw (128.5,285.96) node   [align=center] {{\footnotesize image i}};
\draw (248.75,461) node   [align=center] {{\footnotesize 2D Region Proposal}};
\draw (295, 313) node   [align=center] {{\footnotesize feature maps}};
\draw (295, 345) node   [align=center] {{\footnotesize feature maps}};
\draw (455.75,461) node   [align=center] {{\footnotesize Image to Lidar Association}};
\draw (716.83,389) node   [align=center] {{\footnotesize Frustum PointNets}};
\end{tikzpicture}
\vspace{-1.5em}
\caption{General system overview. The images for all the cameras are processed through the 2D CNN object detector, which proposes 2D regions and
provides the feature maps for each proposal. Those features are then introduced to the siaNMS, which will determine which detections correspond to the same object. Taking this into consideration, the frustum areas for each detection are computed, and the point cloud input for the F-PointNets is calculated. Finally, the 3D box parameters are estimated as proposed by Qi \textit{et al.} in \cite{qi2018frustum}.}
\vspace{-.5em}
\label{fig:general}
\end{figure*}

%Autonomous driving relies on an accurate perception of the environment
Perception for autonomous vehicles is widely dominated by deep learning approaches, which usually process LiDAR and cameras data to estimate the 3D position of the surrounding objects. Within these methods, some are focused on object detection using a single sensor, while others make use of data from multiple modalities.

\textbf{LiDAR 3D detection}. The 3D point cloud captured by laser scanners provides reliable/precise geometry and reflection information in the long-range, typically covering \ang{360} in the horizontal field of view (HFOV). On the contrary, the unstructured nature of LiDAR data caused by its uneven sparsity makes it hard to be processed efficiently. As a result, a discretization of the cloud is often performed before feeding the networks, either in the form of a voxelization \cite{chen2017multi, zhou2018cvpr, yang2019std}, or a Bird Eye's View projection \cite{beltran2018birdnet, yang2018hdnet}.

\textbf{Monocular 3D detection}. Other methods take advantage of the  appearance information provided by camera sensors to perform 3D object detection. Although this kind of modality is well structured and contains rich and dense features, it suffers from the lack of spatial data, a limited HFOV, and weak robustness against light changes. In order to deal with RGB inputs, most approaches propose two-stage solutions \cite{ku2019monocular, brazil2019m3d}. First, 2D proposals are computed using a CNN-based detector, and their estimated depth information is then used to obtain the 3D bounding boxes. However, alternative approaches are presenting networks able to infer the final detections while skipping an explicit prediction of a depth map \cite{bao2019monofenet, simonelli2019single}.

\textbf{Multi-modal fusion 3D detection}. Lately, a range of deep networks that process both camera and LiDAR data has been presented. The main motivation is the possibility to combine complementary information sources to enhance the learned representation of the objects and increase the robustness of the model against adverse conditions. Nevertheless, how these modalities are fused effectively is yet a matter of research. Currently, two distinct lines are followed. On the one hand, a variety of strategies to perform fusion at the feature level \cite{ku2017joint, liang2018deep, liang2019multi} have been introduced. On the other hand, some works divide the process into two steps: performing detection in the image space, and later regressing the 3D box using a subset of the LiDAR modality \cite{qi2018frustum, wang2019frustum}.

Despite the different degrees of maturity of each of these approaches, those using images as input present difficulties when integrated into multiple-cameras setups. First, due to the limited HFOV of these sensors, the accuracy of the detection of objects falling on the side of the image is impaired by truncations. Second, a single object may be detected twice when located within the overlapping area of contiguous cameras.

In this regard, most approaches have opted for greedy NMS algorithms, where all candidate detections are compared to each other to suppress duplicates by computing the IoU of their corresponding boxes and preserving the one with the highest score. However, due to the processing time required to compute the IoU between rotated boxes, an approximation of the method is usually chosen and axis-aligned detections are considered instead, leading to a loss of accuracy.

Alternatively, other methods with similar purposes have been developed within related research fields such as multi-object tracking. For instance, siamese \cite{yi2014deep} or triplet \cite{schroff2015facenet} networks aim to identify multiple occurrences of the same object over time by computing a feature vector in the image space and estimating their similarity by reducing the distance for positive pairs, while increasing it for negative matches.

%%%%%%%%%%%%%%%%%%%%%%%%%%%%%%%%%%%%%%%%%%%%%%%%%%%%%%%%%%%%%%%%%%%%%%%%%%%%%%%%

\section{PROPOSED APPROACH}
%In this paper, we propose a way to re-identify the same obstacle in two contiguous cameras using the previously extracted features from image detection. 
In this paper, we embed a re-identification module into the popular F-PointNets \cite{qi2018frustum} detector, where the camera image is used to obtain object proposals and a further PointNet ensemble estimates the 3D boxes from the corresponding frustum clouds.

Although it is among the state-of-the-art multi-modal fusion approaches, its performance decreases when integrated in multi-camera setups, such as those designed for self-driving perception, as detection of objects truncated by the HFOV of the camera leads to incomplete LiDAR frustums, damaging the final 3D outcome.

To address this issue, a siamese network is used to associate 2D proposals representing the same object in contiguous cameras. A set of fully-connected layers generate an embedding for every proposal inside the overlapping area of the cameras, and a similarity distance is computed. Then, those pairs whose distance falls below a given threshold $\alpha$ are paired. To reduce the processing time of this stage, feature vectors from an intermediate layer of the 2D detector are used as inputs. Finally, matched bounding boxes are used to extract point cloud frustums from LiDAR, which are added together.

%Our approach is divided in three modules: extraction of the feature maps from the image detection network, object re-identification, and frustum obtainment for each one of the detections. Each of this modules is explained bellow. 

An overview of the whole pipeline is shown in Fig.~\ref{fig:general}. As can be seen, the proposed re-identification method is placed between the image detection and the 3D estimation stages.

% TODO Hay que poner algo del párrafo siguiente, pero no se cuánto ni donde
%As our approach pretends to deal with the common areas between cameras, only the detections appearing on the \textit{overlapping region} of each camera will be considered. To calculate this region, the 3D FOV of each camera is obtained, assuming a maximum distance of 50 m, and the overlapping FOV areas are then projected to its image.

\subsection{Feature map extraction from image detection network} \label{sec:approach_feats}
As for the 2D detector, a tuned version of the widespread Faster R-CNN \cite{ren2015faster} framework is selected. For the backbone, a ResNet-50 \cite{he2016deep} with Feature Pyramid Networks (FPN) \cite{lin2017feature} is used. Weights from a model pre-trained on COCO \cite{lin2014microsoft} with an additional instance segmentation branch are used, since image detection benefits from the multi-task learning \cite{he2017mask}.

%{\color{red} implementation from \cite{wu2019detectron2}, more specifically, the pretrained weights from the maskRCNN\_R50\_FPN\_3x configuration}. 

As mentioned before, the siamese network's input vectors are taken from an intermediate feature map generated during the 2D detector inference. Therefore, after the final regression of the 2D boxes is computed by the fully connected (FC) layers of the framework, they are scaled down appropriately to extract their corresponding feature map by applying the ROI Align operation over the output of the fourth ResNet block \textit{res4}. Due to the nature of this pooling layer, an fixed-sized feature vector is obtained for every detection, which can be fed into the re-identification network.
 
\subsection{Object re-identification}
In order to determine which detections correspond to the same obstacle, we consider all detections on the overlapping region of two contiguous cameras as candidates. Then, the object re-identification network processes the extracted features and obtains an embedding for each detection, and the L2 distances between all possible combinations are obtained. If the distance is higher than a defined threshold $dis_{thr}$, the pair is dismissed. The remaining matches are sorted by distance and the ones with the smallest distances are chosen, following the Hungarian Method.

%the embedding for each candidate is calculated 
% to be paired.
%the following steps are taken: %, the distance between the vectors of all possible candidates is calculated and the pair with the minimum distance is chosen.
%for images coming from two contiguous cameras, we consider as candidates to be paired all detections near the common edge between the two cameras. 

%\begin{equation}
%\label{eq:norml2}
%\begin{split}
%distance = \sqrt{i_0^{2}+i_1^{2}+[...]+i_n^{2}}
%\end{split}
%\end{equation}
\comment{
\begin{algorithm}[ht]
\SetAlgoLined
%\KwResult{Write here the result }
 calculate embedding for detections in $image_{i}$\;
 calculate embedding for detections in $image_{j}$\;
 \For{$det_i$ in $image_i$ detections}{
    \For{$det_j$ in $image_j$ detections}{
        $distance_{ij} = \left\|embedding_i - embedding_j\right\|_2$\;
        \If{$distance_{ij} < dis_{thr}$}{
            add $distance_{ij}$ to $all\_distances$\;
        }
    }
 }
 sort $all\_distances$\;
 \For{$distance_{ij}$ in $all\_distances$}{
    create match between i and j\;
    remove all i appearances in $all\_distances$\;
    remove all j appearances in $all\_distances$\;
}
 \caption{Object re-identification algorithm}
 \label{algo}
\end{algorithm}
}

\subsection{Multi-view frustum aggregation}
Finally, to obtain the point cloud inputs to the F-PointNets, the three-dimensional region for each frustum is computed as in \cite{qi2018frustum}. For all the non-paired detections, their frustum is determined from the 2D bounding box and through the camera projection matrix. In the case of paired detections, these steps are followed: first, the frustum region for each detection of the object is calculated; then, if both frustum overlap, the LiDAR points composing the union of both regions are selected as inputs. If they don't, the match is considered a false positive and it is dismissed.

%Finally, in order to obtain the point cloud inputs to the F-PointNets, the three-dimensional region for each frustum is computed. For all the non-paired detections, the 3D region extraction is done as in \cite{qi2018frustum}: from a 2D bounding box and through the camera projection matrix, a frustum  is obtained; all the LiDAR points included in that region are considered inliers. In the case of paired detections, the following steps are followed: first, the frustum region for each detection of the object is calculated. Then, if both frustum regions overlap, the LiDAR points composing the union of both regions are selected as inputs. If they don't, the match is considered a false positive and it is dismissed.

Before feeding frustums into F-Pointnets, they have to be rotated so that their central axis is orthogonal to the image plane. To do so, the center of the 2D bounding box is lifted to its corresponding 3D line, which is used as axis. For multi-view detections, the central axis is calculated as $\overline{or \, p_m}$, where $or$ is the origin of the camera, and $p_m$ is the middle point between $p_l$ and $p_r$, being
%\begin{equation}
%central\_axis = \overline{or \, p_m}
%\label{eq:axis}
%\end{equation}
\begin{equation}
p_l = \mathcal{F}_1 \cap \mathcal{C}_d,
\end{equation}
\begin{equation}
p_r = \mathcal{F}_2 \cap \mathcal{C}_d,
\end{equation}
where $\mathcal{F}_i$ is the frustum region each of the detections and the $\mathcal{C}_d$ is the circumference defined by the maximum detection distance. An example in \textit{Bird's Eye View} can be seen in Fig.~\ref{fig:axis_frustum}.

\definecolor{uququq}{rgb}{0.25,0.25,0.25}
\definecolor{zzccff}{rgb}{0.6,0.8,1}
\definecolor{zzzzff}{rgb}{0.6,0.6,1}
\begin{figure}[h]
\centering
%\vspace{-1em}
\begin{tikzpicture}[line cap=round,line join=round,>=triangle 45,x=0.15cm,y=0.15cm]
\clip(-54,-4.5) rectangle (0.5,17.6);
\fill[fill=black,fill opacity=0.1] (-18.06,2.65) -- (-23.6,-0.31) -- (-25.41,3.08) -- (-19.87,6.04) -- cycle;
\draw (-18.06,2.65)-- (-23.6,-0.31);
\draw (-18.06,2.65)-- (-23.6,-0.31);
\draw (-23.6,-0.31)-- (-25.41,3.08);
\draw (-25.41,3.08)-- (-19.87,6.04);
\draw (-19.87,6.04)-- (-18.06,2.65);
\draw [shift={(-0.49,0.59)},color=zzzzff,fill=zzzzff,fill opacity=0.1]  (0,0) --  plot[domain=2.87:3.12,variable=\t]({1*50.07*cos(\t r)+0*50.07*sin(\t r)},{0*50.07*cos(\t r)+1*50.07*sin(\t r)}) -- cycle ;
\draw [shift={(-0.48,0.1)},color=zzccff,fill=zzccff,fill opacity=0.1]  (0,0) --  plot[domain=2.88:3.16,variable=\t]({1*50.01*cos(\t r)+0*50.01*sin(\t r)},{0*50.01*cos(\t r)+1*50.01*sin(\t r)}) -- cycle ;
\draw [shift={(0,0)}] plot[domain=2.8:3.22,variable=\t]({1*50*cos(\t r)+0*50*sin(\t r)},{0*50*cos(\t r)+1*50*sin(\t r)});
\draw [line width=0.4pt,dash pattern=on 1pt off 1pt] (-47.86,15.06)-- (-50.16,-2.02);
\draw [dash pattern=on 1pt off 1pt on 1pt off 4pt,domain=-53.0:-0.48312803] plot(\x,{(-1.67--6.5*\x)/-48.52});
\draw (-18.96,4.36)-- (-20.9,3.32);
\begin{scriptsize}
\fill [color=uququq,shift={(-0.49,0.59)},rotate=-0] (0,0) ++(0 pt,1.5pt) -- ++(1.3pt,-2.25pt)--++(-2.6pt,0 pt) -- ++(1.3pt,2.25pt);
\draw[color=uququq] (-1.5,2.3) node {$cam_1$};
\fill [color=uququq,shift={(-0.48,0.1)},rotate=-110] (0,0) ++(0 pt,1.5pt) -- ++(1.3pt,-2.25pt)--++(-2.6pt,0 pt) -- ++(1.3pt,2.25pt);
\draw[color=uququq] (-1.5,-1.5) node {$cam_2$};
\fill [color=uququq] (-48.01,13.97) circle (1pt);
\draw[color=uququq] (-50,14.89) node {$p_l$};
\fill [color=uququq] (-49.99,-0.78) circle (1pt);
\draw[color=uququq] (-52,-1.17) node {$p_r$};
\fill [color=uququq] (-49,6.59) circle (1pt);
\draw[color=uququq] (-46, 8) node {$p_m$};
\draw[color=uququq] (-48, -3.5) node {$\mathcal{C}_d$};
\end{scriptsize}
\end{tikzpicture}
\vspace{-1em}
\caption{Multi-frustum axis calculation.}
\label{fig:axis_frustum}
\end{figure}
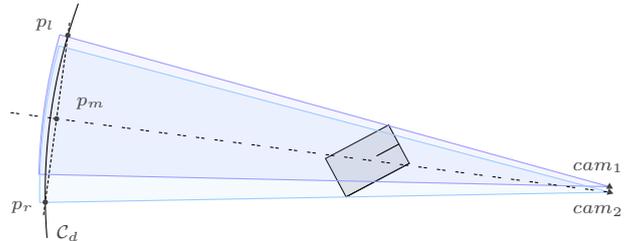

%\begin{figure}[t]
%\centering
%\includegraphics[width=\linewidth]{figures/detectron2_det.png}
%\caption{Detections on Detectron2 of the same pair of images. The detections are made by %inserting the ground truth boxes as proposals for the net. Green detection is shared between %the two cameras. From each detection, a 1x256x7x7 features vector is saved.}
%\label{fig:detectron2_det}
%\end{figure}

%\begin{figure*}[t]
%\centering
%\includegraphics[width=\linewidth]{figures/out.pdf}
%\caption{ESQUEMA GENERAL.}
%\label{fig:general}
%\end{figure*}

%%%%%%%%%%%%%%%%%%%%%%%%%%%%%%%%%%%%%%%%%%%%%%%%%%%%%%%%%%%%%%%%%%%%%%%%%%%%%%%%

\section{Network description and training}
\begin{figure}[!t]
\centering
\includegraphics[width=\linewidth]{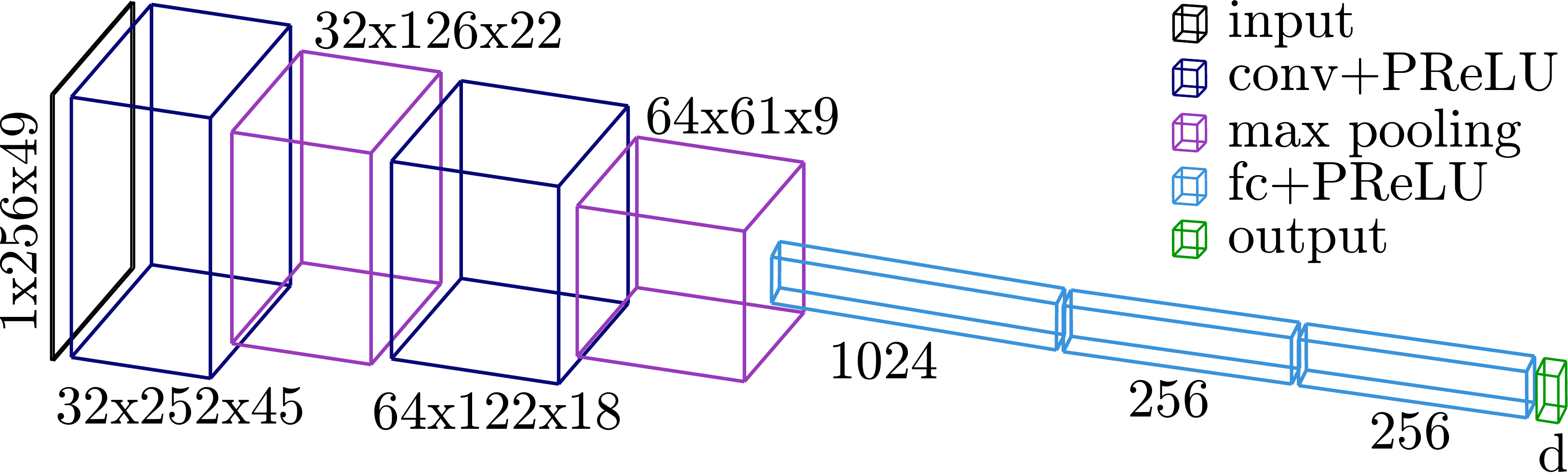}
\caption{Siamese Embedding Network. Each encoder is composed by two groups of convolutional and max pooling layers followed by a set of fully connected layers.}
\label{fig:sianms_net}
\end{figure}

The objective of the siaNMS network is to learn an embedding that transforms the input feature maps into a n-dimensional Euclidean space, and it is trained such that the squared L2 distances between the embeddings of different detections are correlated to the detections similarity. The detailed network architecture is shown in Fig.~\ref{fig:sianms_net}.

\subsection{Loss function definition}
For every input feature map $x$, the output embedding is represented by $f(x) \in \mathbb{R}^{d}$. We want to ensure that the embedding of a specific detection  $f(x^r_i)$ (reference) is closer than a threshold $\alpha$ to the embeddings of all detections $f(x^p_i)$ (positives) of the same object, and that the embedding of any other object $x^n_i$ (negatives) is further away than a threshold $\beta$. To achieve this goal, we have used a Double Margin Contrastive Loss \cite{gomez2018deep}. Thus, we want
\begin{equation}
\left\|f\left(x_{i}^{r}\right)-f\left(x_{i}^{p}\right)\right\|_{2} < \alpha,
\label{eq:margin_a}
\end{equation}
\begin{equation}
\left\|f\left(x_{i}^{r}\right)-f\left(x_{i}^{n}\right)\right\|_{2} > \beta,
\label{eq:margin_b}
\end{equation}
\begin{equation}
\forall\left(f\left(x_{i}^{r}\right), f\left(x_{i}^{p}\right), f\left(x_{i}^{n}\right)\right) \in \mathcal{P}.
\label{eq:forall}
\end{equation}
where $\alpha$ and $\beta$ are two constant margins and $\mathcal{P}$ is the set of all possible image pairs in the training set. The loss that is being minimized is then:
\begin{equation}
\label{eq:triplet_loss}
\begin{split}
        \mathcal{L} = &\frac{1}{2} \sum_{i}^{N}\Big[\max \left( \left\|f\left(x_{i}^{r}\right)-f\left(x_{i}^{p}\right)\right\|_{2}-\alpha, 0 \right)^2 + \\ 
        &\max \left(\beta - \left\|f\left(x_{i}^{r}\right)-f\left(x_{i}^{n}\right)\right\|_{2}, 0 \right)^2\Big].
\end{split}
\end{equation}
%\definecolor{zzqqtt}{rgb}{0.6,0,0.2}
%\definecolor{qqzztt}{rgb}{0,0.6,0.2}
%\begin{figure}
%\begin{tikzpicture}[line cap=round,line join=round,>=triangle 45,x=0.2cm,y=0.2cm]
%\draw[->,ultra thin,color=black] (-6.01,0) -- (9.96,0);
%\foreach \x in {-5,5}
%\draw[shift={(\x,0)},color=black] (0pt,0pt) -- (0pt,-4pt) node[below] {\footnotesize $\x$};
%\draw[->,ultra thin,color=black] (0,-1.49) -- (0,15);
%\foreach \y in {,5,10}
%\draw[shift={(0,\y)},color=black] (2pt,0pt) -- (0pt,0pt) node[right] {\footnotesize $\y$};
%\draw[color=black] (0pt,-10pt) node[right] {\footnotesize $0$};
%\clip(-6.01,-1.49) rectangle (9.96,15);
%\draw[color=qqzztt, smooth,samples=100,domain=1.00:9.96] plot(\x,{1/2*(\x-1)^2});
%\draw[color=qqzztt, smooth,samples=100,domain=-6.00:1.0] plot(\x,{0});
%\draw[color=zzqqtt, smooth,samples=100,domain=3.00:9.96] plot(\x,{0});
%\draw[color=zzqqtt, smooth,samples=100,domain=-6.00:3.0] %plot(\x,{1/2*(3-(\x))^2});
%\begin{scriptsize}
%\draw[color=qqzztt] (4,13) node {t=1};
%\draw[color=zzqqtt] (-4,10.92) node {t = 0};
%\end{scriptsize}
%\end{tikzpicture}
%\caption{Loss shape}
%\label{fig:loss}
%\end{figure}

\subsection{Dataset preparation} 
As the dataset used for assessing the proposed method, nuScenes, is divided into scenes with different driving conditions (e.g. weather, lightning and road kind), an \textit{ad hoc} strategy has been followed to build the training and validation splits. Every scene in the dataset contains a sequence of frames, composed in turn by a set of sensor readings, synchronized by timestamp. Additionally, 3D box labels for objects in the scene are provided, where each object is considered an \textit{instance} with annotations for every occurrence along the set of frames. To train the siamese network, only instances whose 3D box projects inside two consecutive cameras are considered.

In order to prepare the training data, two separate processes are followed. First, an offline computation of the network inputs is performed. Second, an online pairing method is applied during the training phase to select the reference and candidate detections. 

At the offline step, feature maps of these objects have to be obtained so that they can be used as inputs for the re-identification network. As described in \ref{sec:approach_feats}, 2D detection is performed on all the images in the training set. Then, labels are associated with the predicted boxes by computing the IoU in the image plane. To avoid decreasing the training set size, the projection of the 3D ground-truth labels are used for all false negatives.

While training, an Online Hard Example Mining approach \cite{shrivastava2016ohem} is applied in order to form the reference-candidate pairs $\{x_{i}^r, x_{i}^c\}$. Concretely, the following steps are used:
\begin{enumerate}
    \item At every iteration, $N$ reference instances from the training set are taken. For each one, a random occurrence $x_{i}^r$ is picked.
    \item The positive candidate is given by the corresponding bounding box of the same object in the contiguous camera, $x_{i}^p$.
    \item To obtain the negative pair, $x_{i}^n$, another instance appearing in the same scene is randomly selected. Then, every occurrence in the scene of that object in the overlapping area of the contiguous camera is fed into the network, and the corresponding losses are computed. Afterward, the sample with the highest loss value is picked to perform the backward propagation. 
\end{enumerate}

\subsection{Training details}
The proposed network has been trained for 25 epochs with a batch size of 8. An Adam optimizer is used with an initial learning rate of 0.0001, which decays by a factor of 0.1 every 8 epochs. Regarding the loss function parameters, the lower margin value $\alpha = 1$, while the upper threshold $\beta = 3$. Moreover, a $50:50$ ratio of positive and negative samples has been enforced via the online hard negative pair selection method described above.

Besides, the original size of the input images (1600x900) of the 2D detector is scaled down so that the largest size is equal to 1333px. No data augmentation techniques are used.

Different output dimensions for the object embeddings
($d = \left\{5, 10, 20, 50, 100, 200, 500, 1000\right\}$) have been tested. According to our validation tests, the 100-dimensional encoding provides the best performance. This can be explained because the amount of parameters is correlated to the capability of representing the appearance of an object. However, from a certain amount on, the network starts to overfit and is unable to generalize properly for examples not included in the training set.

%%%%%%%%%%%%%%%%%%%%%%%%%%%%%%%%%%%%%%%%%%%%%%%%%%%%%%%%%%%%%%%%%%%%%%%%%%%%%%%%

\section{RESULTS}

The presented approach is evaluated using nuScenes detection benchmark \cite{nuscenes}. Concretely, our method falls within the \textit{Open Track} due to the use of both camera and LiDAR information as input. Since the main contribution of this paper is the introduction of a re-identification module in the perception pipeline as an alternative to traditional NMS, instead of the ten classes included in the official detection challenge, the three classes (Car, Pedestrian, and Cyclist) from the original model in F-PointNets paper are used.

%\begin{figure*}[!ht]
%    \centering
%    \subfloat[]{
%        \centering
%        \includegraphics[width=0.33\linewidth]{figures/normal/car_pr.png}
        %\includegraphics[width=0.47\linewidth]{figures/car_tp_normal.png}
%        \label{fig:car_normal}
%    }\hspace{-1em}\subfloat[]{
%        \centering
%        \includegraphics[width=0.33\linewidth]{figures/car_pr_axis_nms.png}
        %\includegraphics[width=0.47\linewidth]{figures/pedestrian_tp_normal.png}
%        \label{fig:car_axis_nms}
%    }\hspace{-1em}\subfloat[]{
%        \centering
%        \includegraphics[width=0.33\linewidth]{figures/joined/car_pr.png}
        %\includegraphics[width=0.47\linewidth]{figures/bicycle_tp_normal.png}
%        \label{fig:car_joined}
%    }
%    \caption{Evaluation results nuScenes val split. Precision-recall curves for Car class. (a) Baseline, (b) Axis NMS, (c) siaNMS.}
%    \label{fig:curves-normal}
%\end{figure*}
\comment{
\begin{figure*}
\centering
%\missingfigure{Figure is missing}
\includegraphics[width=0.45\linewidth]{figures/im_view2879.png}
\includegraphics[width=0.45\linewidth]{figures/im_view2879.png}
\\
\includegraphics[width=0.45\linewidth]{figures/im_view2879.png}
\includegraphics[width=0.45\linewidth]{figures/im_view2879.png}
\caption{Ground-truth 3D boxes on the NuScenes dataset. Single image detections are colored in red. Meanwhile, detection shared between two cameras are colored blue.}
\end{figure*}
}

\begin{figure*}
\captionsetup[subfloat]{farskip=2pt,captionskip=1pt}
\subfloat{
    \\
    \includegraphics[width=0.1644\linewidth]{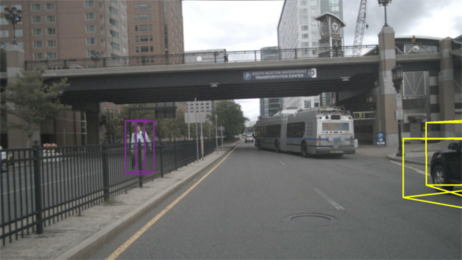}\\
    \includegraphics[width=0.1644\linewidth]{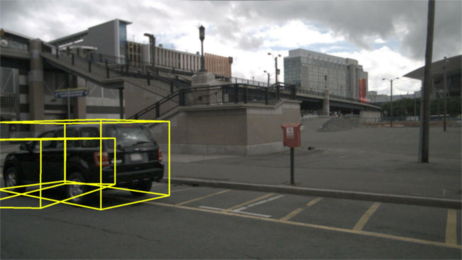}
    \includegraphics[width=0.1644\linewidth]{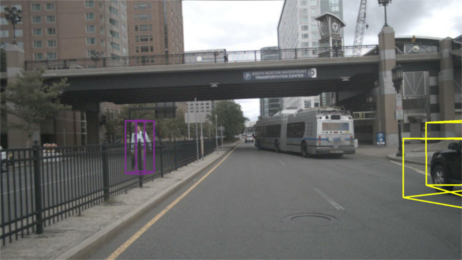}\\
    \includegraphics[width=0.1644\linewidth]{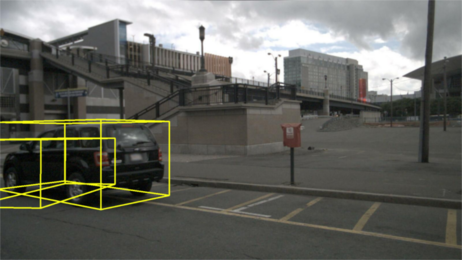}
    \includegraphics[width=0.1644\linewidth]{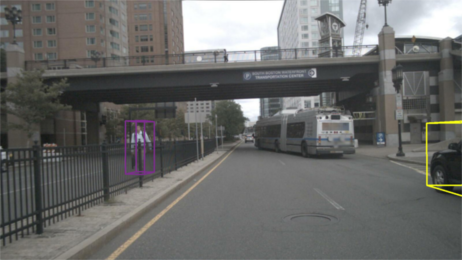}\\
    \includegraphics[width=0.1644\linewidth]{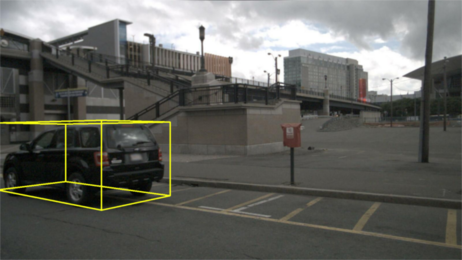}
} \\
\subfloat{
    \\
    \includegraphics[width=0.1644\linewidth]{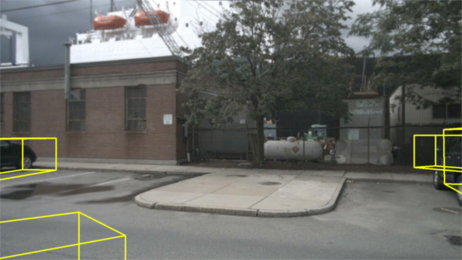}\\
    \includegraphics[width=0.1644\linewidth]{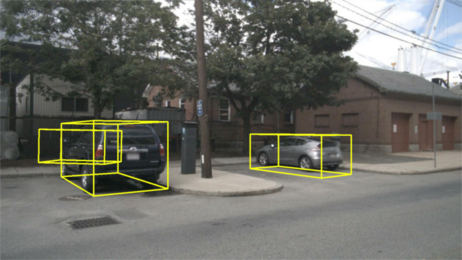}
    \includegraphics[width=0.1644\linewidth]{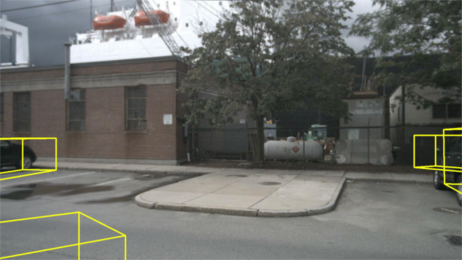}\\
    \includegraphics[width=0.1644\linewidth]{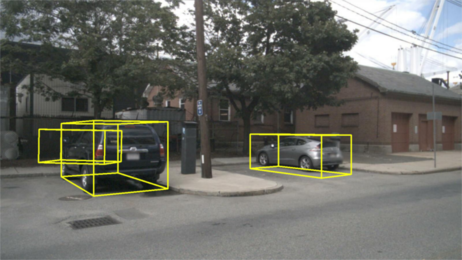}
    \includegraphics[width=0.1644\linewidth]{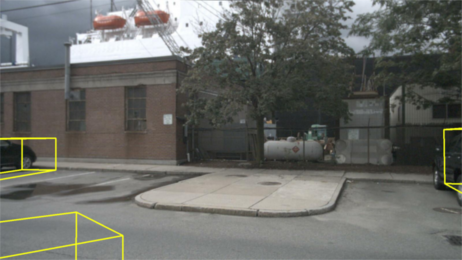}\\
    \includegraphics[width=0.1644\linewidth]{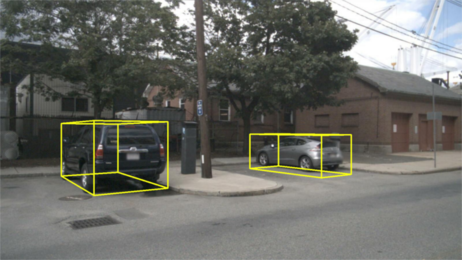}
}\\
\subfloat{
    \\
    \includegraphics[width=0.1644\linewidth]{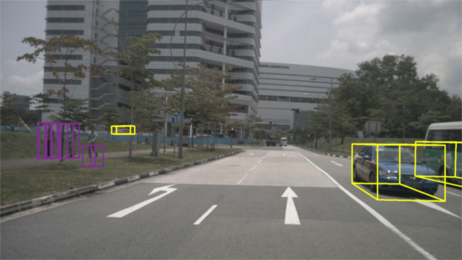}\\
    \includegraphics[width=0.1644\linewidth]{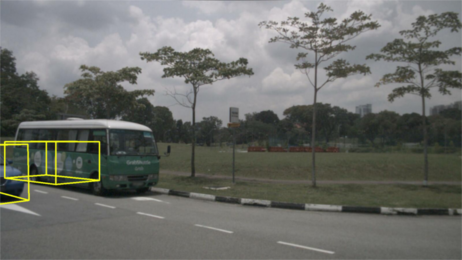}
    \includegraphics[width=0.1644\linewidth]{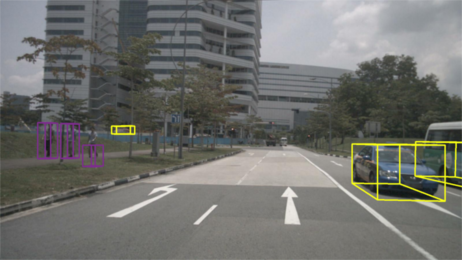}\\
    \includegraphics[width=0.1644\linewidth]{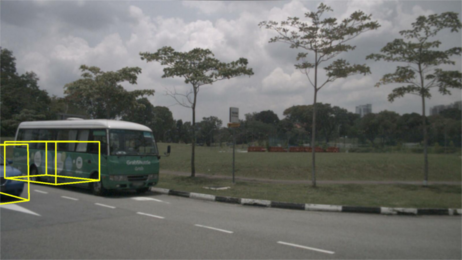}
    \includegraphics[width=0.1644\linewidth]{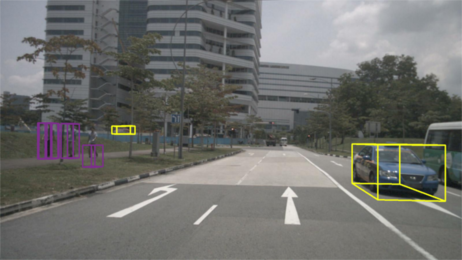}\\
    \includegraphics[width=0.1644\linewidth]{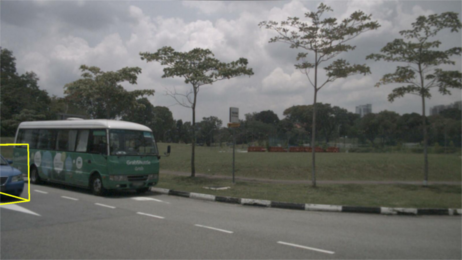}
}\\
\subfloat{
    \\
    \includegraphics[width=0.1644\linewidth]{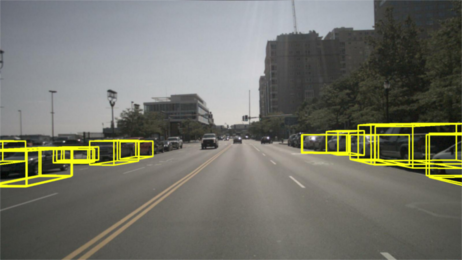}\\
    \includegraphics[width=0.1644\linewidth]{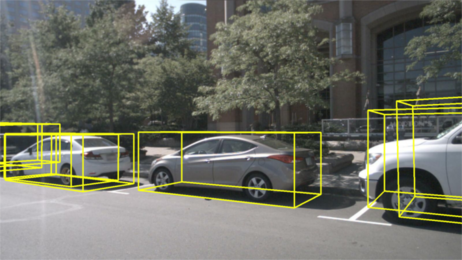}
    \includegraphics[width=0.1644\linewidth]{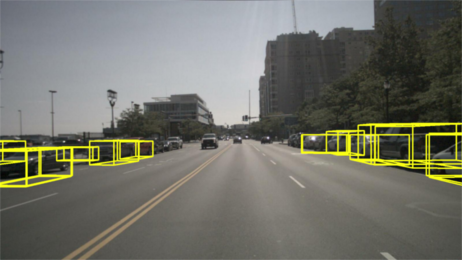}\\
    \includegraphics[width=0.1644\linewidth]{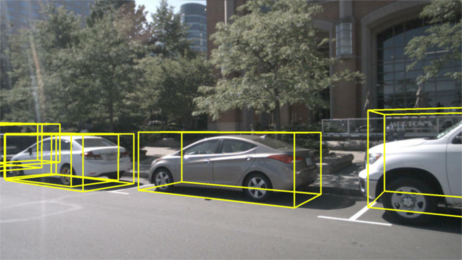}
    \includegraphics[width=0.1644\linewidth]{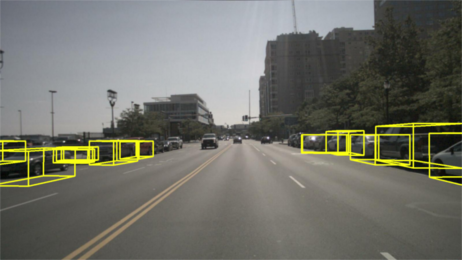}\\
    \includegraphics[width=0.1644\linewidth]{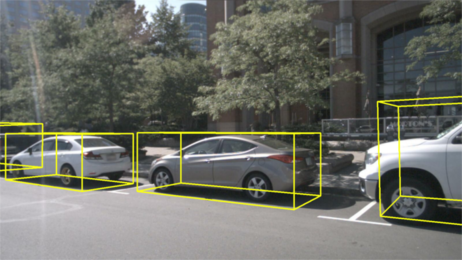}
}\\
\subfloat{
    \\
    \includegraphics[width=0.1644\linewidth]{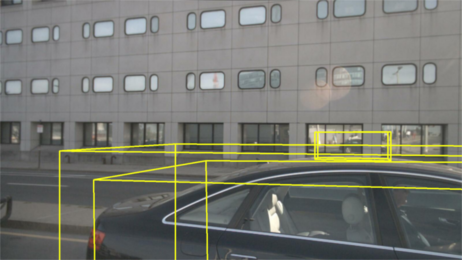}\\
    \includegraphics[width=0.1644\linewidth]{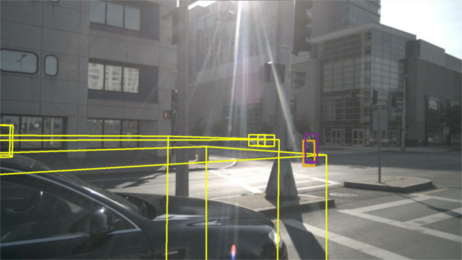}
    \includegraphics[width=0.1644\linewidth]{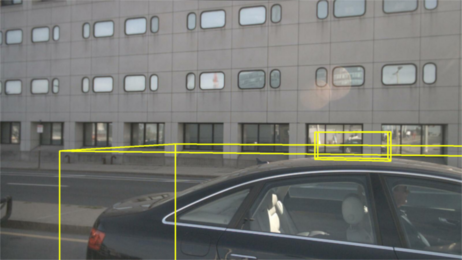}\\
    \includegraphics[width=0.1644\linewidth]{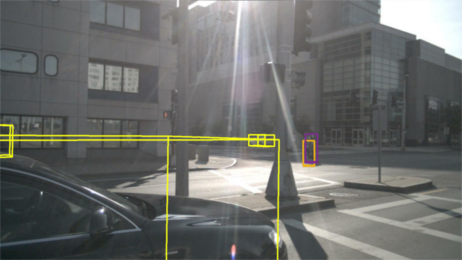}
    \includegraphics[width=0.1644\linewidth]{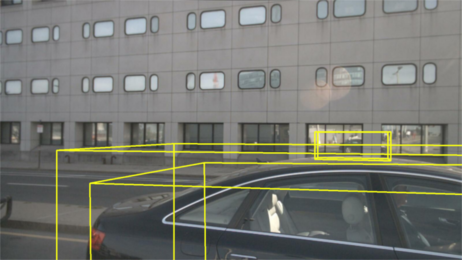}\\
    \includegraphics[width=0.1644\linewidth]{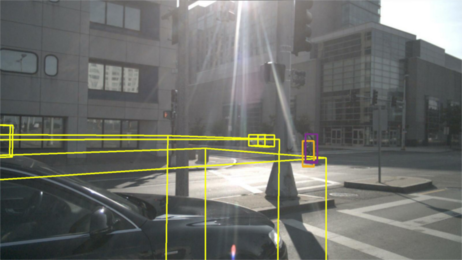}
}
\comment{\\
\subfloat{
    \\
    \includegraphics[width=0.1644\linewidth]{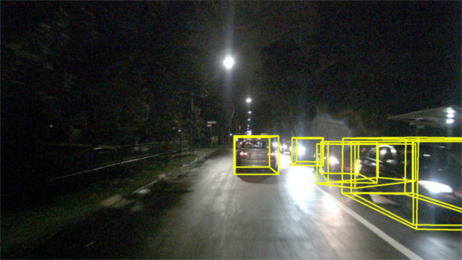}\\
    \includegraphics[width=0.1644\linewidth]{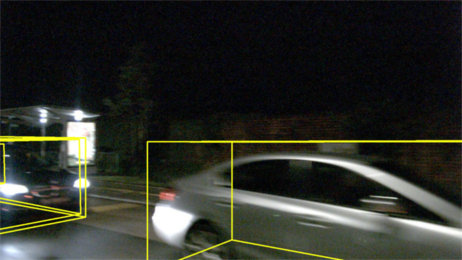}
    \includegraphics[width=0.1644\linewidth]{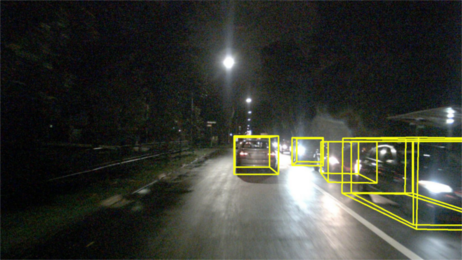}\\
    \includegraphics[width=0.1644\linewidth]{fotos/im_nms03view2107000019.png}
    \includegraphics[width=0.1644\linewidth]{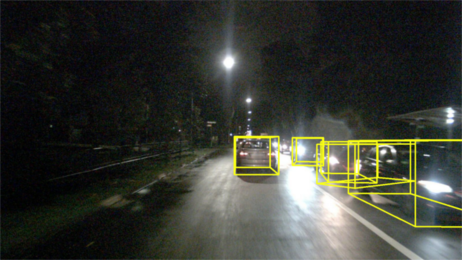}\\
    \includegraphics[width=0.1644\linewidth]{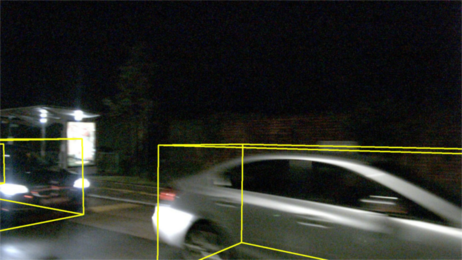}
}}
\caption{Results on nuScenes validation set. From left to right: 3D detections using Vanilla, Axis-NMS and siaNMS approaches.}
\label{fig:examples}
\end{figure*}

\begin{table*}[ht]
\caption{Comparison of the 3D Car Detection Performance on the nuScenes Validation Set in Different Regions of Interest}
\vspace{-1em}
\label{tab:table-car}
\centering
\begin{tabular}{l cccc cccc cccc}
\toprule
Areas & \multicolumn{4}{c}{Vanilla} & \multicolumn{4}{c}{Axis NMS} & \multicolumn{4}{c}{siaNMS} \\
\cmidrule(l){2-5} \cmidrule(l){6-9} \cmidrule(l){10-13} 
   & AP & ATE & ASE & AOE & AP & ATE & ASE & AOE & AP & ATE & ASE & AOE  \\
 \midrule
 All & 45.5 & 0.331 & 18.4 & 0.371 & 49.1 & 0.330 & 18.4 & 0.367 & \textbf{51.1} & \textbf{0.320} & \textbf{18.2} & \textbf{0.350} \\
Overlap & 37.9 & 0.317 & 18.6 & 0.312 & 46.6 & 0.312 & 18.6 & 0.306 & \textbf{49.0} & \textbf{0.287} & \textbf{18.0} & \textbf{0.253}\\

 \bottomrule
\end{tabular}
\end{table*}

\subsection{Evaluation metrics}
The used metrics are defined in the nuScenes detection benchmark \cite{nuscenes}:
\begin{itemize}
    \item Average Precision (AP) [\%]: a true positive (TP) is defined if the 2D center distance between a detection and a label is smaller than a threshold. This metric is calculated for thresholds of $\{0.5, 1, 2, 4\}$ meters, and then the average for each class is calculated.
\end{itemize}
The remaining metrics are only calculated for TP detections:
\begin{itemize}
    \item Average Translation Error (ATE) [m]: Average distance between detection and label centers.
    \item Average Scale Error (ASE) [\%]: The IoU after aligning centers and orientation between detection and label 3D boxes is calculated. ASE is defined as 1 - IoU.
    \item Average Orientation Error (AOE) [rad]: Yaw angle difference between detection and label boxes in radians.
\end{itemize}

Other metrics %defined in the nuScenes benchmark 
such as Average Velocity Error (AVE) and Average Attribute Error (AAE) are not taken into account as they do not apply to the purpose of this paper.

\subsection{Experimental setup}
To evaluate the performance of the proposed approach, results over the nuScenes validation set are provided. For a fair analysis, we consider the same pipeline for 3D detections only with variations in the method to address the grouping of detections between cameras.
As shown in Table~\ref{tab:table-car}, a comparison between three approaches has been considered:
\begin{enumerate}
    \item A vanilla version of the F-PointNets is used for every camera. Afterward, all the object detections are transformed to the global frame.
    \item Detections from the previous solution are filtered using a traditional Greedy NMS to suppress duplicates, with an IoU threshold of 0.3. An axis-aligned NMS approach in Bird's Eye View is selected over a rotated one due to its suitability for real-time applications.
    \item The proposed siamese network is used to remove multiple detections of the same object in contiguous cameras. This process is only applied to cars, as they are by far the most frequently truncated class among the considered ones, due to its dimensions.
\end{enumerate}

To better understand the impact of the proposed siamese network, the same evaluation is performed taking into account only the regions of overlap between cameras.

\subsection{Discussion}

As can be seen in the results shown in Table~\ref{tab:table-car}, the presented approach outperforms traditional NMS in all evaluated metrics even though the input feature maps have not been optimized for re-identification tasks. The one that benefited the most is the AP since a large number of redundant objects are removed. Nonetheless, all other metrics are also improved to some extent, due to the fact that having more complete point clouds for truncated obstacles results advantageous for the quality of the 3D box regression. All these effects are magnified when the analysis is performed isolating the overlap areas, as there is were most duplicate detections are found. %Such is the case of the ATE metric, where center misalignment error is reduced by more than 20cm with respect to the other methods.

As shown in Fig.~\ref{fig:examples}, cases where either the object is truncated in both images or the detection on one of the cameras leads to a wrong 3D box are better resolved by the siaNMS. In such situations, the 3D estimations usually present a greater misalignment due to the incomplete LiDAR input, and may not have sufficient overlap for the NMS to be able to suppress it. These examples are the most benefited from the multi-view frustum aggregation feature of the proposed approach, as can be observed in the first four rows of Fig.~\ref{fig:examples}.

Despite the above, we have detected cases in which the re-identification method works worse than the traditional NMS, see last row of Fig.~\ref{fig:examples}. These are cases in which the object appears with very different perspectives in both images, e.g. when the obstacle is very close to the camera, or the light conditions are not adequate, such as reflections, overexposure, etc. Hence, both methods might be used together as they can provide a complementary behavior in edge case scenarios.

%%%%%%%%%%%%%%%%%%%%%%%%%%%%%%%%%%%%%%%%%%%%%%%%%%%%%%%%%%%%%%%%%%%%%%%%%%%%%%%%

\section{CONCLUSIONS}
In this paper, an effective alternative to NMS for suppressing duplicate detections of the same object in multi-camera setups has been presented. To this end, a siamese network has been added between the detection and the 3D box regression stages of a top-performing 3D object detector. Moreover, the association of 2D detection boxes has been exploited to obtain a more reliable representation of the objects in the LiDAR space, improving the quality of the inputs of the subsequent stage in the pipeline.

The proposed work has been evaluated in a challenging \ang{360} object detection benchmark, proving its capability to cope with complex %driving
scenarios. According to the experimental results, the embedded re-identification network outperforms the traditional NMS method in average precision and reduces the translation, size and orientation errors thank to the aggregation of 3D frustums from matched image detections.

In future work, several models will be trained to permit the re-identification of other kinds of classes, paying special attention to bulky objects, which are more prone to be truncated by camera images. Besides, the siamese layers will be integrated into the perception pipeline creating an end-to-end deep neural network. Thus, following a  multi-task learning strategy, the encoder will be able to compute feature maps that are best suited for both 2D detection and re-identification purposes.

%%%%%%%%%%%%%%%%%%%%%%%%%%%%%%%%%%%%%%%%%%%%%%%%%%%%%%%%%%%%%%%%%%%%%%%%%%%%%%%%

\section*{ ACKNOWLEDGMENT} % TODO: TO BE REVIEWED!
Research supported by the Spanish Government through the CICYT projects (TRA2016-78886-C3-1-R and RTI2018-096036-B-C21), Universidad Carlos III of Madrid (PEAVAUTO-CM-UC3M) and the Comunidad de Madrid (SEGVAUTO-4.0-CM P2018/EMT-4362). We gratefully acknowledge the support of NVIDIA Corporation with the donation of the GPUs used for this research.

%%%%%%%%%%%%%%%%%%%%%%%%%%%%%%%%%%%%%%%%%%%%%%%%%%%%%%%%%%%%%%%%%%%%%%%%%%%%%%%%

\bibliographystyle{IEEEtran}
\bibliography{paper}

\begin{thebibliography}{10}
\providecommand{\url}[1]{#1}
\csname url@samestyle\endcsname
\providecommand{\newblock}{\relax}
\providecommand{\bibinfo}[2]{#2}
\providecommand{\BIBentrySTDinterwordspacing}{\spaceskip=0pt\relax}
\providecommand{\BIBentryALTinterwordstretchfactor}{4}
\providecommand{\BIBentryALTinterwordspacing}{\spaceskip=\fontdimen2\font plus
\BIBentryALTinterwordstretchfactor\fontdimen3\font minus
  \fontdimen4\font\relax}
\providecommand{\BIBforeignlanguage}[2]{{%
\expandafter\ifx\csname l@#1\endcsname\relax
\typeout{** WARNING: IEEEtran.bst: No hyphenation pattern has been}%
\typeout{** loaded for the language `#1'. Using the pattern for}%
\typeout{** the default language instead.}%
\else
\language=\csname l@#1\endcsname
\fi
#2}}
\providecommand{\BIBdecl}{\relax}
\BIBdecl

\bibitem{kitti_object}
A.~Geiger, P.~Lenz, and R.~Urtasun, ``Are we ready for autonomous driving? the
  kitti vision benchmark suite,'' in \emph{Conference on Computer Vision and
  Pattern Recognition (CVPR)}, 2012.

\bibitem{nuscenes}
H.~Caesar, V.~Bankiti, A.~H. Lang, S.~Vora, V.~E. Liong, Q.~Xu, A.~Krishnan,
  Y.~Pan, G.~Baldan, and O.~Beijbom, ``{nuScenes}: A multimodal dataset for
  autonomous driving,'' \emph{arXiv preprint arXiv:1903.11027}, 2019.

\bibitem{waymo}
P.~Sun, H.~Kretzschmar, X.~Dotiwalla, A.~Chouard, V.~Patnaik, P.~Tsui, J.~Guo,
  Y.~Zhou, Y.~Chai, B.~Caine \emph{et~al.}, ``Scalability in perception for
  autonomous driving: An open dataset benchmark,'' \emph{arXiv preprint
  arXiv:1912.04838}, 2019.

\bibitem{lyft2019}
R.~Kesten, M.~Usman, J.~Houston, T.~Pandya, K.~Nadhamuni, A.~Ferreira, M.~Yuan,
  B.~Low, A.~Jain, P.~Ondruska, S.~Omari, S.~Shah, A.~Kulkarni, A.~Kazakova,
  C.~Tao, L.~Platinsky, W.~Jiang, and V.~Shet, ``Lyft level 5 av dataset
  2019,'' url{https://level5.lyft.com/dataset/}, 2019.

\bibitem{argoverse}
M.-F. Chang, J.~Lambert, P.~Sangkloy, J.~Singh, S.~Bak, A.~Hartnett, D.~Wang,
  P.~Carr, S.~Lucey, D.~Ramanan, and J.~Hays, ``Argoverse: 3d tracking and
  forecasting with rich maps,'' in \emph{The IEEE Conference on Computer Vision
  and Pattern Recognition (CVPR)}, June 2019.

\bibitem{qi2018frustum}
C.~R. Qi, W.~Liu, C.~Wu, H.~Su, and L.~J. Guibas, ``Frustum pointnets for 3d
  object detection from rgb-d data,'' in \emph{Proc. IEEE Conference on
  Computer Vision and Pattern Recognition (CVPR)}, 2018, pp. 918--927.

\bibitem{chen2017multi}
X.~Chen, H.~Ma, J.~Wan, B.~Li, and T.~Xia, ``Multi-view 3d object detection
  network for autonomous driving,'' in \emph{Proceedings of the IEEE Conference
  on Computer Vision and Pattern Recognition}, 2017, pp. 1907--1915.

\bibitem{zhou2018cvpr}
Y.~Zhou and O.~Tuzel, ``Voxelnet: End-to-end learning for point cloud based 3d
  object detection,'' in \emph{The IEEE Conference on Computer Vision and
  Pattern Recognition (CVPR)}, June 2018.

\bibitem{yang2019std}
Z.~Yang, Y.~Sun, S.~Liu, X.~Shen, and J.~Jia, ``Std: Sparse-to-dense 3d object
  detector for point cloud,'' in \emph{Proceedings of the IEEE International
  Conference on Computer Vision}, 2019, pp. 1951--1960.

\bibitem{beltran2018birdnet}
J.~Beltran, C.~Guindel, F.~M. Moreno, D.~Cruzado, F.~Garcia, and
  A.~De~La~Escalera, ``Birdnet: a 3d object detection framework from lidar
  information,'' in \emph{2018 21st International Conference on Intelligent
  Transportation Systems (ITSC)}.\hskip 1em plus 0.5em minus 0.4em\relax IEEE,
  2018, pp. 3517--3523.

\bibitem{yang2018hdnet}
B.~Yang, M.~Liang, and R.~Urtasun, ``Hdnet: Exploiting hd maps for 3d object
  detection,'' in \emph{Conference on Robot Learning}, 2018, pp. 146--155.

\bibitem{ku2019monocular}
J.~Ku, A.~D. Pon, and S.~L. Waslander, ``Monocular 3d object detection
  leveraging accurate proposals and shape reconstruction,'' in
  \emph{Proceedings of the IEEE Conference on Computer Vision and Pattern
  Recognition}, 2019, pp. 11\,867--11\,876.

\bibitem{brazil2019m3d}
G.~Brazil and X.~Liu, ``M3d-rpn: Monocular 3d region proposal network for
  object detection,'' in \emph{Proceedings of the IEEE International Conference
  on Computer Vision}, 2019, pp. 9287--9296.

\bibitem{bao2019monofenet}
W.~Bao, B.~Xu, and Z.~Chen, ``Monofenet: Monocular 3d object detection with
  feature enhancement networks,'' \emph{IEEE Transactions on Image Processing},
  2019.

\bibitem{simonelli2019single}
A.~Simonelli, S.~R. Bul{\`o}, L.~Porzi, E.~Ricci, and P.~Kontschieder,
  ``Single-stage monocular 3d object detection with virtual cameras,''
  \emph{arXiv preprint arXiv:1912.08035}, 2019.

\bibitem{ku2017joint}
J.~Ku, M.~Mozifian, J.~Lee, A.~Harakeh, and S.~Waslander, ``Joint 3d proposal
  generation and object detection from view aggregation,'' \emph{arXiv preprint
  arXiv:1712.02294}, 2017.

\bibitem{liang2018deep}
M.~Liang, B.~Yang, S.~Wang, and R.~Urtasun, ``Deep continuous fusion for
  multi-sensor 3d object detection,'' in \emph{Proceedings of the European
  Conference on Computer Vision (ECCV)}, 2018, pp. 641--656.

\bibitem{liang2019multi}
M.~Liang, B.~Yang, Y.~Chen, R.~Hu, and R.~Urtasun, ``Multi-task multi-sensor
  fusion for 3d object detection,'' in \emph{Proceedings of the IEEE Conference
  on Computer Vision and Pattern Recognition}, 2019, pp. 7345--7353.

\bibitem{wang2019frustum}
Z.~Wang and K.~Jia, ``Frustum convnet: Sliding frustums to aggregate local
  point-wise features for amodal 3d object detection,'' in \emph{IROS}.\hskip
  1em plus 0.5em minus 0.4em\relax IEEE, 2019.

\bibitem{yi2014deep}
D.~Yi, Z.~Lei, S.~Liao, and S.~Z. Li, ``Deep metric learning for person
  re-identification,'' in \emph{2014 22nd International Conference on Pattern
  Recognition}.\hskip 1em plus 0.5em minus 0.4em\relax IEEE, 2014, pp. 34--39.

\bibitem{schroff2015facenet}
F.~Schroff, D.~Kalenichenko, and J.~Philbin, ``Facenet: A unified embedding for
  face recognition and clustering,'' in \emph{Proceedings of the IEEE
  conference on computer vision and pattern recognition}, 2015, pp. 815--823.

\bibitem{ren2015faster}
S.~Ren, K.~He, R.~Girshick, and J.~Sun, ``Faster r-cnn: Towards real-time
  object detection with region proposal networks,'' in \emph{Advances in neural
  information processing systems}, 2015, pp. 91--99.

\bibitem{he2016deep}
K.~He, X.~Zhang, S.~Ren, and J.~Sun, ``Deep residual learning for image
  recognition,'' in \emph{Proceedings of the IEEE conference on computer vision
  and pattern recognition}, 2016, pp. 770--778.

\bibitem{lin2017feature}
T.-Y. Lin, P.~Doll{\'a}r, R.~Girshick, K.~He, B.~Hariharan, and S.~Belongie,
  ``Feature pyramid networks for object detection,'' in \emph{Proceedings of
  the IEEE conference on computer vision and pattern recognition}, 2017, pp.
  2117--2125.

\bibitem{lin2014microsoft}
T.-Y. Lin, M.~Maire, S.~Belongie, J.~Hays, P.~Perona, D.~Ramanan,
  P.~Doll{\'a}r, and C.~L. Zitnick, ``Microsoft coco: Common objects in
  context,'' in \emph{European conference on computer vision}.\hskip 1em plus
  0.5em minus 0.4em\relax Springer, 2014, pp. 740--755.

\bibitem{he2017mask}
K.~He, G.~Gkioxari, P.~Doll{\'a}r, and R.~Girshick, ``Mask r-cnn,'' in
  \emph{Proceedings of the IEEE international conference on computer vision},
  2017, pp. 2961--2969.

\bibitem{gomez2018deep}
M.~J. G{\'o}mez-Silva, J.~M. Armingol, and A.~de~la Escalera, ``Deep parts
  similarity learning for person re-identification.'' in \emph{VISIGRAPP (5:
  VISAPP)}, 2018, pp. 419--428.

\bibitem{shrivastava2016ohem}
A.~Shrivastava, A.~Gupta, and R.~Girshick, ``Training region-based object
  detectors with online hard example mining,'' in \emph{Proceedings of the IEEE
  conference on computer vision and pattern recognition}, 2016, pp. 761--769.

\end{thebibliography}

\end{document}